\documentclass[runningheads]{llncs}

\usepackage{graphicx}

\graphicspath{ {./images/} }
\begin{document}
\title{ConveRT for FAQ Answering}
%
%
\author{Maxime De Bruyn \and
Ehsan Lotfi \and
Jeska Buhmann \and
Walter Daelemans}
\authorrunning{M. De Bruyn et al.}
%
\institute{CLiPS Research Center\\
Antwerp University, Belgium\\
\email{firstname.lastname@uantwerpen.be}}
\maketitle              
\begin{abstract}
Knowledgeable FAQ chatbots are a valuable resource to any organization. While powerful and efficient retrieval-based models exist for English, it is rarely the case for other languages for which the same amount of training data is not available. In this paper, we propose a novel pre-training procedure to adapt ConveRT, an English conversational retriever model, to other languages with less training data available. We apply it for the first time to the task of Dutch FAQ answering related to the COVID-19 vaccine. We show it performs better than an open-source alternative in both a low-data regime and a high-data regime.

\end{abstract}

\section{Introduction}

In this paper, we present a Dutch-based FAQ retrieval system trained using a limited amount of training data.

FAQ answering is the task of retrieving the right answer given a new user query. It is widely used in chatbots and has been studied for many years \cite{hammond1995faq,sneiders1999automated,jijkoun2005retrieving,riezler2007statistical,karan2016faqir,sakata2019faq}, although the attention has shifted towards extractive question answering more recently \cite{rogers2021qa}, probably because of a lack of dedicated datasets. FAQ answering systems typically use retrieval systems \cite{hammond1995faq,sneiders1999automated,jijkoun2005retrieving,riezler2007statistical,karan2016faqir,sakata2019faq} rather than generative models grounded on external knowledge \cite{komeili2021internet,de2020bart,lotfi2021teach}. The generative approach is more flexible as it is able to generate new answers. However, these models suffer from knowledge hallucinations \cite{shuster2021retrieval}, limiting their usefulness in a corporate environment.

Most previous research focusing on FAQ retrieval and non-factoid question answering were developed for English. ConveRT \cite{DBLP:journals/corr/abs-1911-03688}, a response selection module available within Rasa \cite{DBLP:journals/corr/abs-1712-05181}, caught our attention as it is effective and does not require a GPU at inference time. Unfortunately, it is only available in English. Despite having significantly less conversational training data (400K pairs of utterances) than the original ConveRT model (727M pairs), we successfully trained the same model for Dutch.

Our contributions are the following:
\begin{itemize}
\item We show it is possible to train a ConveRT model for a non-English language using a limited number of conversation pairs by adopting a two-phase pre-training approach (general and conversational).
\item We show that a Dutch ConveRT model performs better than the response selector module from Rasa, both in a low and high data regime.
\end{itemize}

\section{Related Work}
An FAQ dataset consists of pairs of questions and answers. The FAQ retrieval task involves ranking the available answers for a given user query. There are three methods available to solve this problem: matching a new user query on the available questions, the answers, or the concatenation of both. FAQ retrieval can be broadly divided into 4 categories: lexical, supervised, unsupervised, and conversational.

\paragraph{Lexical}
To our knowledge, FAQ-Finder \cite{hammond1995faq} was the first to explicitly study the task of FAQ retrieval, it tries to do so by matching user queries to FAQ questions of the Usenet dataset with TF-IDF. FAQ-Finder was later improved by including the similarity to the answer (on top of the similarity to the question) \cite{tomuro2004retrieval}. Another improvement comes from adding a rule-based layer on top of the TF-IDF module \cite{sneiders1999automated}.

\paragraph{Unsupervised}
Another approach is to used unsupervised techniques to retrieve the right FAQ pair given a new user query. One possible way is to use Latent Semantic Analysis (LSA) to overcome the lexical mismatch between related queries \cite{kim2008cluster}. 

\paragraph{Supervised}
The first supervised methods were developed using tree kernels and SVMs \cite{moschitti2007exploiting}. BERT methods were later developed specifically for the task of FAQ retrieval \cite{sakata2019faq}.

\paragraph{Conversational}
In this paper, we propose a fourth type not yet explored in the literature: conversational. FAQ retrieval can be treated as a special case of conversational modeling: retrieving the answer is similar to retrieving the next utterance in a conversation.

Dual-encoder architectures, pre-trained on response selection, have become increasingly popular in the dialog community due to their simplicity and ease of control \cite{DBLP:journals/corr/abs-1906-01543,cer-etal-2018-universal}.
There are two options when it comes to retrieving the next utterance. One can either encode the two sentences separately (dual-encoder) \cite{DBLP:journals/corr/abs-1911-03688}, or simultaneously (cross-encoder) \cite{10.1007/978-3-030-47426-3_19}. Dual-encoders are faster than cross-encoders as they can cache the answer representations. ConveRT \cite{DBLP:journals/corr/abs-1911-03688} is a dual-encoder pre-trained on a large-scale conversational dataset. Thanks to various design optimizations (such as using single-headed self-attention) ConveRT can vastly reduce the size of the model.

In this work, we choose to focus on ConveRT as it has a low computational cost and does not require a GPU for inference.

\section{ConveRT}
In this section, we give a brief overview of the ConveRT (Conversational Representations from Transformers) model \cite{DBLP:journals/corr/abs-1911-03688}. The objective of the model is to generate vector representations for utterances that are as similar as possible (in terms of dot-product) for a given pair. ConveRT takes as input the sequence of tokens of the two utterances. Both sequences are tokenized using the same byte pair encoding vocabulary.

\subsection{Architecture}
\begin{figure}[t]
\includegraphics[scale=0.45]{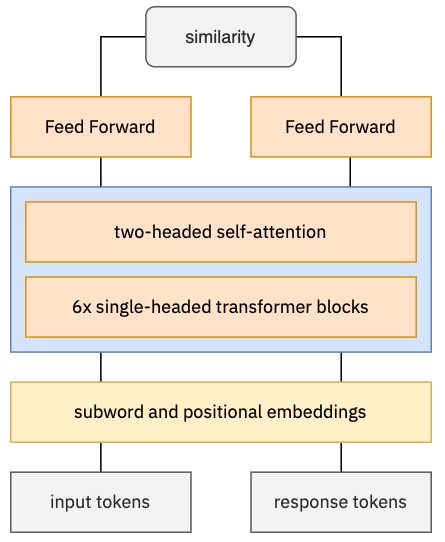}
\centering
\caption{Illustration of the ConveRT model architecture. The model has three distinct parts. First, the subword and positional embeddings. Second, a shared Transformer block followed by a two-headed self-attention. Third, separate feed-forward networks (3 layers) for the input and responses.}
\label{figure:architecture}
\end{figure}
The ConveRT architecture (Fig. \ref{figure:architecture}) is composed of three distinct parts: the embedding layer, the Transformer block and the feedforward layers.

\subsubsection{Embedding}
The first element stores the embeddings for the subwords and position tokens. Embeddings are shared for the input and response representations. Unlike the original Transformer architecture \cite{DBLP:journals/corr/VaswaniSPUJGKP17}, ConveRT uses two positional encoding matrices of different sizes to handle sequences larger than seen during training. We refer the reader to the original paper for a detailed description \cite{DBLP:journals/corr/abs-1911-03688}.

\subsubsection{Transformer Block}
The next element is the Transformer block. It closely follows the original Transformer architecture \cite{DBLP:journals/corr/VaswaniSPUJGKP17} with some notable differences.
First, the model uses a single-headed self-attention using a 64-dimensional projection for computing the attention weights. Second, the model applies a two-headed self-attention after the six Transformer layers. 
The parameters of the Transformer block are fully shared for the input and response sides. ConveRT uses the square-root-of-N reduction \cite{cer-etal-2018-universal} to convert the embedding sequences to fixed-dimensional vectors.

\subsubsection{Feed Forward}
The last elements are a series of feed-forward hidden layers with skip connections. The parameters are not shared between the inputs and responses side, as there is a separate feed-forward for the inputs and responses.

\subsection{Training Objective}
The training objective of ConveRT is to select the right response given a question from a question-answer pair.
The relevance of each response to a given input is quantified with a dot-product between the input and response representation.
Training proceeds in a batch of K pairs of utterances. The objective is to distinguish between the true relevant responses and irrelevant negative examples (we use other responses from the batch as negative examples). ConveRT uses cross-entropy as the loss function. The model is optimized with Adam \cite{DBLP:journals/corr/KingmaB14} and L2 weight decay. The learning rate is warmed up over the first 10,000 steps to a peak value and then linearly decayed.

\section{ConveRT for Dutch}
In this section, we explain our approach to training a ConveRT model for Dutch.
To overcome the limited supply of conversational data available in Dutch, we use a two-stage pre-training: general pre-training on a large open-domain corpus, and conversational pre-training using a smaller conversational dataset from Reddit.

\subsection{Data}
The original ConveRT model was developed for English using a large-scale conversational dataset from Reddit.
We did not have access to such a dataset for Dutch. Instead, we chose to split the problem in two. First, we pre-train the model on a general Dutch corpus. Second, we use a smaller Dutch conversational corpus from Reddit. 

\subsubsection{General Dataset}
We consider the same Dutch-language corpora as Bertje \cite{devries2019bertje}, a successful Dutch BERT model:
\begin{itemize}
\item Books: a collection of contemporary and historical fiction novels
\item TwNC \cite{42e3c5016cab421281a9029a774fffae}: a Multifaceted Dutch News Corpus
\item SoNaR-500 \cite{Oostdijk2013}: a multi-genre reference corpus
\item Web news
\item Wikipedia
\end{itemize}
In total, this is about 12GB of uncompressed text.

To match the setup expected by ConveRT (the tokens of a pair of utterances), we first split each paragraph into sentences. Next, we save pairs of sentences and treat them as pairs of input and response. To avoid small inputs, we filter out pairs with less than 64 characters. After transformation, the general corpus dataset for pre-training has 110M pairs.

\subsubsection{Conversational Dataset}
We also consider a Dutch conversational dataset for which we downloaded comments from around 200 Dutch subreddits. Non-Dutch comments were filtered out. After filtering for the language we arrive at a size of 400K pairs of utterances.

\subsection{Pre-training}
We followed the training procedure of ConveRT, except for the number of epochs and the batch size. For the general pre-training, we trained the model for 8 epochs. To facilitate the training, we used other examples from the batch as negative examples.

To increase the difficulty of the training, we doubled the batch size at every second epoch. The batch size increased from 128 at the first epoch to 2048 at the last epoch. The larger the batch size, the harder it is for the model as the model has to select the correct response amongst more negative examples.

For the conversational pre-training, we trained for 10 epochs with a fixed batch size of 2048.

\begin{table*}
\centering
\begin{tabular}{lcccccc}
\hline
\textbf{model}                       & \textbf{split 1} & \textbf{split 2} & \textbf{split 4} & \textbf{split 6} & \textbf{split 8} & \textbf{split 10} \\ \hline
RASA (baseline)                                & 22\%             & 42\%             & 50\%             & 55\%             & 61\%             & 65\%              \\
without pre-training                      & 20\%             & 25\%             & 33\%             & 45\%             & 52\%             & 65\%              \\
general pre-training                  & 30\%             & 36\%             & 40\%             & 55\%             & 58\%             & 43\%              \\
conversational pre-training           & 40\%             & 44\%             & 55\%             & 63\%             & 66\%             & 69\%              \\
general + conversational pre-training & \textbf{46\%}    & \textbf{57\%}    & \textbf{68\%}    & \textbf{69\%}    & \textbf{75\%}    & \textbf{79\%}     \\ \hline
\end{tabular}
\caption{
Accuracy on the COVID-19 vaccination FAQ dataset per splits of increasing size. Split one has one training example per answer, while split ten has ten training examples. Pre-training ConveRT on both a general dataset, as well as a conversational dataset provides the best results on this task.
}
\label{table:results}
\end{table*}

\section{Experiments}
In this section, we fine-tune our model on a corpus of FAQs related to the COVID-19 vaccine. We then perform an ablation study to analyze which part of the pre-training has the most impact on the downstream performance. To have a better understanding of how our model would perform in the real world, we study its performance as the number of training examples increases.

\subsection{Data}
We test the performance of our model on a proprietary dataset. The dataset was collected while running a COVID-19 vaccination FAQ bot with Rasa. It consists of 1,200 questions for 76 distinct answers.

\subsection{Baseline}
As our higher objective is to use this model in a Rasa chatbot, we compare our Dutch ConveRT model to a baseline response retrieval model developed by Rasa.\footnote{Rasa does not have a published paper describing their model.} All models are trained using the same number of epochs and dropout probability.

\subsection{Low Data Scenario}
When starting out, FAQ bots usually have a one-on-one mapping between the number of questions and answers (one question for one answer). As the number of users increases, the number of available questions per answer also increases. To evaluate the generalization capabilities of our model in a low data scenario, we artificially create datasets of increasing sizes, which we call splits. The first split has one training example per answer (the same as when someone starts a new FAQ chatbot), the second split has two training examples per answer, and so on until split ten. We also generate a test set by randomly selecting (and removing from the training set) one training example per answer.

\subsection{Results}

Results in Table \ref{table:results} confirm our intuition that the baseline accuracy of the Rasa model radically improves with the number of training examples. In our analysis, the accuracy increases by a factor of 3 from split 1 to split 10. The results also show that a ConveRT model without any pre-training  underperforms the baseline, on every split. General pre-training modestly improves the model's performance, but the results are not significantly different from the baseline. Conversational pre-training alone (without any general pre-training) shows a consistent improvement over the baseline. The gain is more visible in the low data regime than in the high data regime. The Dutch ConveRT model reveals its true power when pre-trained on a general corpus and a conversational corpus as it outperforms the baseline by a wide margin on every split.

\section{Conclusion}
We have successfully pre-trained, fine-tuned, and evaluated a Dutch ConveRT model. This model consistently outperforms a baseline response selector from Rasa on a COVID-19 vaccine FAQ dataset. 

Conversational datasets for non-English languages are scarce. Our two-phase pre-training procedure bypasses this problem by first pre-training on a general corpus, then pre-training on a smaller conversational corpus.

In future work, we plan on extending the two-stage training to additional languages and additional domains.

\section{Acknowledgments}
This research received funding from the Flemish Government under the “Onderzoeksprogramma Artificiële Intelligentie (AI) Vlaanderen” programme. We also thank the reviewers for their helpful comments.





\bibliographystyle{plain}
\bibliography{anthology,acl2021}
\end{document}